\documentclass[letterpaper, 10 pt, conference]{ieeeconf}  

\IEEEoverridecommandlockouts                              

\usepackage{graphics} 
\usepackage{epsfig} 
\usepackage{amsmath} 
\usepackage{amssymb}  
\usepackage{amsfonts}
\usepackage{multirow}
\usepackage{booktabs}
\usepackage{multirow}
\usepackage{bm}
\usepackage{subfigure}
\usepackage{cite}

\title{\LARGE \bf
Enhance Planning with Physics-informed Safety Controller for End-to-end Autonomous Driving}

\author{Hang Zhou$^{1}$$^{,}$$^{3}$, Haichao Liu$^{1}$$^{,}$$^{4}$, Hongliang Lu$^{2}$, Dan Xu$^{3}$, Jun Ma$^{1}$$^{,}$$^{4}$ and Yiding Ji$^{1}$$^{,}$$^{4}$
\thanks{$^{1}$Robotics and Autonomous Systems Thrust, Systems Hub,
            The Hong Kong University of Science and Technology (Guangzhou), Guangzhou, China
        {\tt\small hzhou269@connect.hkust-gz.edu.cn}; {\tt\small hliu369@connect.hkust-gz.edu.cn}; {\tt\small jun.ma@ust.hk}; {\tt\small jiyiding@hkust-gz.edu.cn}}%
\thanks{$^{2}$Intelligent Transpotation Thrust, Systems Hub,
            The Hong Kong University of Science and Technology (Guangzhou), Guangzhou, China
        {\tt\small hlu592@connect.hkust-gz.edu.cn}}
\thanks{$^{3}$Department of Computer Science and Engineering, School of Engineering, The Hong Kong University of Science and Technology, Hong Kong SAR, China
        {\tt\small  danxu@cse.ust.hk}} 
\thanks{$^{4}$Department of Electronic and Computer Engineering, School of Engineering, The Hong Kong University of Science and Technology, Hong Kong SAR, China}
}

\begin{document}

\maketitle
\thispagestyle{empty}
\pagestyle{empty}

\begin{abstract}
Recent years have seen a growing research interest in applications of Deep Neural Networks (DNN) on autonomous vehicle technology. The trend started with perception and prediction a few years ago and it is gradually being applied to motion planning tasks. Despite the performance of networks improve over time, DNN planners inherit the natural drawbacks of Deep Learning. Learning-based planners have limitations in achieving perfect accuracy on the training dataset and network performance can be affected by out-of-distribution problem. In this paper, we propose FusionAssurance, a novel trajectory-based end-to-end driving fusion framework which combines physics-informed control for safety assurance. By incorporating Potential Field into Model Predictive Control, FusionAssurance is capable of navigating through scenarios that are not included in the training dataset and scenarios where neural network fail to generalize. The effectiveness of the approach is demonstrated by extensive experiments under various scenarios on the CARLA benchmark.


\end{abstract}

\section{Introduction}
Autonomous Vehicle Technology (AVT) is a rapidly developing field that aims to facilitate safe and efficient navigation through complex and dynamic traffic environments. One of the key challenges in this area is to design intelligent systems that can perceive, predict, and plan for various driving scenarios based on sensory data from cameras, radar, lidar and other devices. Deep Neural Network (DNN) is a powerful class of machine learning model that can learn high-level feature representations from large-scale data, and has demonstrated remarkable performance in autonomous driving domains such as perception, prediction and planning. 

\begin{figure}[t]
\centering
\includegraphics[ width=1.0\linewidth]{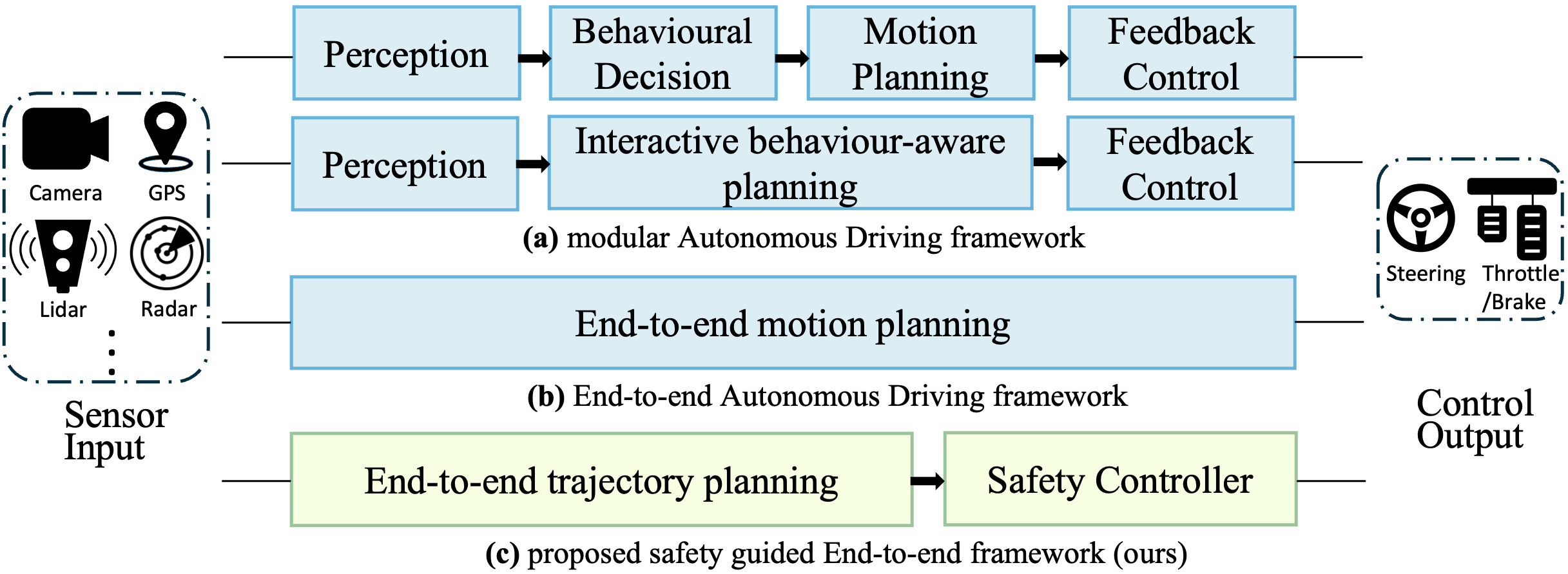}
\caption{Popular frameworks of Autonomous Driving~\cite{schwarting2018planning}. Green shaded is the proposed framework pipeline of FusionAssurance. Compared to other frameworks, FusionAssurance is less modular and its DNN output can be guided by safety controller.}
\label{fig:pipeline}
\end{figure}

The modular AVT architecture in Fig.\ref{fig:pipeline}(a) is explainable~\cite{7410669} and entails mass deployment~\cite{Yurtsever_2020}. However, recent trends started to recast modular AVT architecture into a single network. End-to-end autonomous driving in Fig.\ref{fig:pipeline}(b) can potentially learn more complex and nuanced behaviors that are difficult to capture by hand-crafted rules or algorithms in modular systems. For example, end-to-end systems learn how to handle rare or unexpected situations such as road hazards, pedestrians and traffic violations, by observing human drivers and mimicking their actions. End-to-end autonomous driving reduces the computational cost and latency of processing the sensor input. This improves the responsiveness and safety of the vehicle, especially in high-speed or dynamic scenarios. End-to-end autonomous driving also avoids error propagation and inconsistency issues that may arise from integrating multiple modules with different assumptions and representations~\cite{Tampuu_2022}.

However, DNN-based planner does not guarantee reliability, safety and optimality due to out-of-distribution~\cite{hendrycks17baseline} and generalization issue. Since learning-based motion planner has limitations in achieving perfect accuracy, it also has natural drawback generalizing existing known scenarios. On the other hand, out-of-distribution problem is a challenge in machine learning that occurs when the test data differs from the training data in some way. This can lead to poor performance, overconfidence or incorrect predictions by the machine learning model. Real-world driving scenarios are dynamic, complex and therefore often unpredictable. The performance of data-centric motion planner is contingent on the quality and quantity of the training data. The network may not perform well in situations that are absent from the training data or simulations.

To address these issues, we propose an innovative mapless trajectory-based end-to-end driving framework which combines Model Predictive Control and Potential Field for safety assurance. The main contributions are as follows:
\begin{itemize} 
\item The proposed framework enables integrated decision-making and control of lane keeping, adaptive cruise control and overtaking for data-centric motion planners.
\item The proposed method allows the accommodation of diverse driving behaviors by tuning parameters in the safety control module without the need to retrain the entire neural network.
\item The proposed framework allows agents to navigate through unseen dynamic and complex scenarios where DNN planner failed to generate feasible trajectories. Our method surpasses preceding methodologies and has the best performance on CARLA 42 routes benchmark.
\end{itemize}

\section{Related Work}

\textbf{End-to-end Autonomous Driving} can be generally categorised as trajectory-based method and direct control-based method. Trajectory-based models are usually combined with a controller (e.g. PID) to execute the planned motion. Direct control-based network has its control actions directly optimised as network output. Although the two approaches differ in their methodologies, they can benefit from each other in the aspects of feature extraction and sensor fusion.

Direct control-based end-to-end models are usually achieved by reinforcement Learning (RL) ~\cite{9562006,chen2021gri,chen2021learning} or imitation learning (IL)~\cite{9009463, cil, chen2019lbc,mile2022,zhang2021endtoend}. Latent DRL~\cite{9562006} creates a latent representation as an embedding space and applies RL to learn the latent observation. GRI~\cite{chen2021gri} presents RL algorithm which efficiently both expert demonstration and environment exploration under predicted maps and traffic condition. CIL~\cite{cil} and CILRS~\cite{9009463} are early IL methods that use a conditional architecture with different network branches for different navigation commands. LBC~\cite{chen2019lbc} and Roach~\cite{zhang2021endtoend} train agents with privileged information and later imitation learn the privileged agents. MILE~\cite{mile2022} applies Model-based Imitation Learning technique which learns a driving policy, mapping, detection and traffic rules in bird's eye view (BEV) concurrently.

Transfuser~\cite{Chitta2022PAMI,Prakash2021CVPR} has a trajectory-based end-to-end multi-modal transformer to learn surrounding scene understanding in terms of depth estimation, road segmentation, map and obstacle detection. It has a PID controller with formal rules for safety mechanism. LAV~\cite{chen2022lav} applies technique of recording trajectory history of all surrounding vehicles and the ego vehicle, which significantly enlarges the diversity and quantity of the training dataset.
IVMP~\cite{9561334} proposed an end-to-end motion planning approach which deploys network-predicted semantic map for better uncertain object handling and trajectory prediction. To take advantage of both trajectory-based and control-based networks, TCP~\cite{wu2022trajectoryguided} designs a single network that produces control signal and waypoints simultaneously. An adaptive ensemble network is used to merge the two outputs for refinement. 

Trajectory-based approaches have the advantage of producing more interpretable result for neural network output than control-based approaches, but they also have the drawback of controller error which means that the actual motion may deviate from the desired trajectory~\cite{https://doi.org/10.4218/etrij.15.0114.0123}.
 In comparison, our approach offers the benefit of interpretability and overcomes the drawback of trajectory-based methods by converting the control tracking error into intelligent decision-making and control with the use of safety controller.

\begin{figure*}[t]
\centering
\includegraphics[ width=1.0\linewidth]{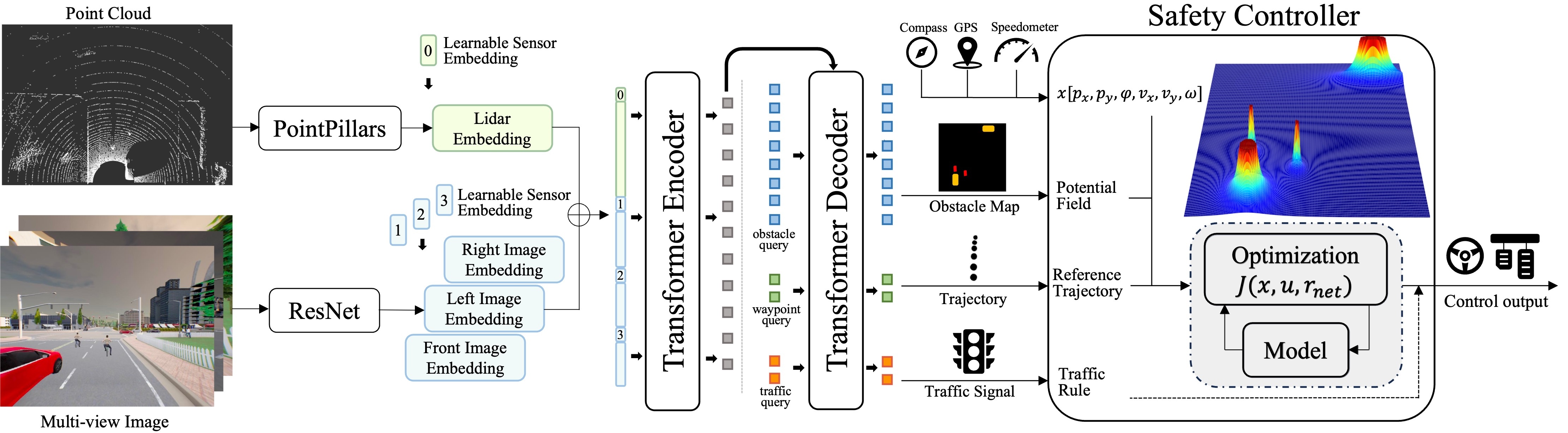}
\caption{Overall structure of FussionAssurance consists of two parts. 1) a transformer-based network that integrates the target location, multi-camera data, and lidar data to generate the predicted waypoint, BEV obstacle map, traffic light state, and junction probability for autonomous driving. 2) a safety controller that comprises Model Predictive Controller with Potential Functions. The safety controller takes the output of the neural network and sensor data of GPS, Compass and Speedometer to produce safe and optimal low-level control actions.}
\label{fig:overall_pipeline}
\end{figure*}

\textbf{Safety controller} has been well studied in modern control theory~\cite{8796030,1087247} and applications.
Rasekhipour et al.~\cite{7576661} proposed a MPC which models road boundary, crossable and non-crossable obstacles as Potential Function to find optimal path and perform obstacle avoidance in predefined scenarios.
Liu et al.~\cite{liu2023integrated} further implemented a Potential Field based MPC on the CARLA simulator. The controller takes the ground truth localization, map and perception information as input and incorporates the traffic control signal as potential functions to achieve integrated decision-making and control. 

Safety in learning-based planning remains to be a critical problem. Wei et al.~\cite{9561904} proposed attention-based safety mechanism in neural network by gating backbone features for direct motion optimisation and coupling safe actions with perception. Kalaria et al.~\cite{kalaria2023safety} presented a RL method that leverages IL and control barrier functions to ensure safety.
InterFuser~\cite{shao2022interfuser} brought up the concept of safety controller for end-to-end autonomous driving. Like other trajectory-based methods, InterFuser uses transformer to decode feature from multi-sensors and output obstacles and waypoints. InterFuser further deploys a simple safety controller to keep distance with target objects which lie in its predicted waypoints. In contrast, our approach incorporates all detected obstacles as input and determines the optimal control output through a global optimization process.

One other related work is DIPP~\cite{huang2023differentiable}. An end-to-end planning network is trained under the guidance of a MPC controller to learn better from its expert dataset. Our approach also use MPC which serves as additional safety layer for neural network. Users can tune the safety controller parameter for different driving behavior without retrain the entire network. Since FusionAssurance applies to any neural network planner, DIPP can also adapt our framework to provide extra safety for its neural network. 

\section{Methodology}
%

The general structure of the proposed safe autonomous driving framework consists of two parts. The first is a transformer-based network for perception and planning. The second part is physics-informed safety controller which consists of Model Predictive Controller with Potential Field Functions. Safety controller takes the output of neural transformer and generates safe low-level control actions.

\subsection{Network Model} 

The proposed neural network follows DETR’s~\cite{carion2020end} general structure to fuse multiple sensor data. Specifically, Transformer is used to fuse three RGB cameras' input ($I_{front}$, $I_{left}$, $I_{right}$) and Lidar sensor input ($I_{lidar}$) together. Each source is followed up with backbone to extract complex features and then fed into Transformer Encoder for feature fusion in BEV. The learned feature is then decoded into waypoints, BEV obstacle map, probability of junction and traffic control signal. 

Camera streams take pretrained ResNet~\cite{he2016deep} as backbone to convert images to feature embedding. Feature resolution changes to $\frac{H}{8}*\frac{W}{8}*256$ from original image resolution $H*W*3$.  Lidar stream on the other hand has different data representation and PointPillars~\cite{lang2019pointpillars} is applied for feature extraction and data resolution reduction. Lidar data consists of three-dimensional (x, y, z) points that are stored without any rule-based order, which implies that adjacent points in visualization may have their data located far apart in the point cloud. Therefore, point cloud data requires preprocessing before being fed into a convolutional neural network. Specifically, a simplified PointNet is employed to transform 3D lidar data into 2D pillar format. Point cloud data therefore can be fed into standard 2D Convolutional Neural Network. Lidar stream network takes point cloud ranged $[-l_y, 0, l_y, l_x]$ as input and assigns it into pillars sized $0.25m * 0.25m$. The feature resolution would be $8l_y * 4l_x*64$ after PointNet and $l_y * \frac{l_x}{2}*512$ after 3-layer CNN.

Following the implementation of ViLT~\cite{kim2021vilt}, single-modal spatial features are linearized and attached with modal-type embedding. Each modal embedding is then concatenated into a one-dimensional token embedding for sensor fusion feature learning. Sinusoidal positional encoding is used to capture the sequential or spatial relationships between feature tokens. Standard Visual Transformer~\cite{dosovitskiy2020vit} encoder with self-attention layer and feedforward layer are used. The self-attention layer enables the encoder to learn the relationships among the sensor tokens in the input sequence, by computing a weighted average of all the tokens based on their similarity. The feedforward layer applies a non-linear transformation to each token individually to enhance its representation. The output of each sub-layer is normalized and added to the sub-layer input to preserve the information from the previous layer. The Transformer has global attention for multiple modalities and captures the global context of the scene.

The decoder receives the tokens from the RGB images as values and keys, and the tokens from the LiDAR points as queries of size $H * W$, which are used to produce features in bird's eye view. Moreover, the decoder also takes two other types of queries for predicting traffic signals and waypoints $w$. Following the approach of InterFuser, a two-layer multilayer perceptron (MLP) is deployed as the traffic sign classifier, which determines the state of the traffic light and the presence of a stop sign ahead. A single-layer gated recurrent unit (GRU)~\cite{cho2014learning} is used to generate successive waypoints $w_{t_s}$ in an auto-regressive manner, which conditioned on the goal location of the ego vehicle. Where ${t_s}$ denotes the number of the predicted time steps.

\subsection{Physics-informed Safety Controller}

The proposed physic-informed controller is based on Model Predictive Control and Potential Field Function. The safety controller takes the output of transformer and decodes them into trajectory waypoints, obstacle  map, probability of junction, red light and stop sign.

\subsubsection{Vehicle Dynamics}
In order to fully utilise the potential of MPC, we leverage a simple but accurate model for the vehicle dynamics which was proposed in~\cite{9669260}. 
\begin{align}
\label{bicycle_dynamics_equation}
\begin{split}
    & \bm x_{k+1}=f(\bm x_{k}, \bm u_{k})\\
    & =\left[\begin{array}{c}
    p_x(k)+\left(v_x(k) \cos \varphi(k)-v_y(k) \sin \varphi(k)\right)\Delta t \\
    p_y(k)+\left(v_y(k) \cos \varphi(k)+v_x(k) \sin \varphi(k)\right)\Delta t \\
    \varphi(k)+\omega(k)\Delta t \\
    v_x(k)+a(k)\Delta t \\
    \frac{m v_x(k) v_y(k)+l \omega(k)\Delta t-k_f \delta(k) v_x(k)\Delta t-m v_x(k)^2 \omega(k)\Delta t}{m v_x(k)-\left(k_f+k_r\right)\Delta t} \\
    \frac{I_z v_x(k) \omega(t)+l v_y(k)\Delta t- l_f k_f \delta(k) v_x(k)\Delta t}{I_z v_x(k)-\left(l_f^2 k_f+l_r^2 k_r\right)\Delta t}
    \end{array}\right]
\end{split}
\end{align}

The above model demonstrates high numerical stability and robustness in urban driving task scenarios which require frequent changes of velocity. The state vector of the dynamics model is $\bm x = [p_x, p_y, \varphi, v_x, v_y, \omega]^T$, $[p_x, p_y]^T$ is the vehicle position in the global map coordinate, and  $[v_x, v_y]^T$ denotes the velocity in the ego vehicle coordinate. Moreover, $\phi$ indicates the ego vehicle orientation with respect to the map coordinate, and $\omega$ indicates the yaw rate. The control input of the system is $\bm u = [a,\delta]^T$ , where $a$ and $\delta$ indicate the acceleration and steering angle respectively. $m$ is the vehicle mass, $l_f$ and $l_r$ represent the distance from the mass center to the front and rear axle, $k_f$ and $k_r$ to denote the cornering stiffness of the front and rear wheels, and $I_z$ is the inertia polar moment. We define $l=l_f k_f - l_r k_r$ for simplicity.

\subsubsection{Potential Field Function}
The cost function of the MPC incorporates repulsive potential field functions for the detected objects, which enables obstacle avoidance capability. In order to avoid surrounding obstacles smoothly and computationally efficiently, elliptic functions are used to describe the PF of obstacles where obstacles' length and width are linear to ellipse vertex and co-vertex. Having a higher obstacle PF value lengthwise helps ego vehicle to perform obstacle avoidance and ACC, and relative smaller PF value crosswise helps close fleets which move in similar orientations not to interfere with each other. The main traffic participants on the road are vehicles, cyclists and pedestrians. Cyclists and pedestrians have smaller physical dimensions but higher vulnerability, therefore bigger PF gain ${K_o}$ is assigned to these two classes. 
\begin{equation}\label{Vehicle_PF}
    F_{O} = \sum_{i=0}^n \dfrac{K_o}{\dfrac{(p_{x_{rot}}-p_{i,x})^2}{a^2}+\dfrac{(p_{y_{rot}}-p_{i,y})^2}{b^2}}
\end{equation}
\begin{equation}
\setlength{\arraycolsep}{1pt}
\bigl[
\begin{array}{c}
p_{x_{rot}} \\
p_{y_{rot}}
\end{array}
\bigr]= 
\\
\bigl[
\begin{array}{c}
p_{i,x} \\
p_{i,y}
\end{array}
\bigr]
\\+\,
\bigl[
\begin{array}{cc}
p_x - p_{i,x}&0 \\
0&p_y - p_{i,y}
\end{array}
\bigr]
\bigl[
\begin{array}{cc}
cos(\theta)&sin(\theta) \\
-sin(\theta)&cos(\theta)
\end{array}
\bigr]
\end{equation}

 where $K_{o}$ controls the intensity of the obstacle repulsive force, which varies under different scenarios. $a$ and $b$ are the half-length of the major and minor axis, which defines the shape of the ellipse. Specific values of obstacle's length and width are assigned to $a$ and $b$ to represent the physical size of the obstacle. $p_{i,x}$ and $p_{i,y}$ denote $i$th obstacle's position. $\theta$ is the orientation of the target obstacle. $p_{x_{rot}}$ and $p_{y_{rot}}$ are the rotated position of the ego vehicle to align with the ellipse.

An extra potential function is applied to the target front obstacle to enhance Adaptive Cruise Control capability. The problem is simplified only to consider the euclidean distance between the ego vehicle and the target front vehicle. 
\begin{equation}\label{Fc}
    F_{C} = \dfrac{K_c * u_{k[0]} * x_{k[2]}}{D_{safety} + 0.001}
\end{equation}
where $K_c$ denotes the gain of the cost function, $u_{k[0]}$ is the throttle control and $x_{k[2]}$ is the current speed of the ego vehicle. $D_{safety}$ denotes the euclidean distance between the ego vehicle and obstacle which lies in its moving trajectory. The cost function penalizes acceleration and high speed when there are obstacles in the front. By having the front obstacle cost function, it helps agent to slow down without trying to perform unnecessary overtaking actions for most simple scenarios.

\subsubsection{Model Predictive Control Formulation}
The cost function is formalised to follow reference trajectory, reach target velocity and improve driving comfort in the receding horizon manner. 
\begin{align}
\label{basic_cost}
\begin{aligned}
    J(\bm x, \bm u, \bm r_{net}) &= \sum_{k=0}^N \| \boldsymbol x_{\text{ref},k}- \boldsymbol x_k \| ^2 + \sum_{k=1}^N \|\bm u_{k}-\bm u_{k-1}\|^2 \\&+\sum_{k=0}^N \|\boldsymbol u_{k}\|^2  
    +F_O(\bm r_{net},\bm x) + F_C(\bm r_{net},\bm x)
\end{aligned}
\end{align}
where $\bm r_{net}$ is the transformed output of the neural network, $N$ is the length of the receding horizon, $\bm x_{\text{ref},k}\in \mathbb{R}^{6}$ is the reference trajectory point generated by spline interpolation based on the network waypoint prediction at the time step $k$. $F_O(\bm r_{net},\bm x)$ and  $F_C(\bm r_{net},\bm x)$ are the artificial potential field functions generated according to the perception information. 
Therefore, the MPC controller is constructed as:
\begin{equation}
\label{NMPCOptProb}
\begin{array}{ll}
\underset{{\boldsymbol{x}}, {\boldsymbol{u}}}{\min} & J(\bm x, \bm u, \bm r_{net})\\
\text { s.t.} & \boldsymbol x_{k+1}=f(\boldsymbol x_k,\boldsymbol u_k), \forall k \in \{0,1,...,N\} \\
& -\boldsymbol u_\text{min} \preceq \boldsymbol u_k \preceq \boldsymbol u_\text{max}, \forall k \in \{0,1,...,N\} \\
& -\boldsymbol x_\text{min} \preceq \boldsymbol x_k \preceq \boldsymbol x_\text{max}, \forall k \in \{0,1,...,N\}
\end{array}
\end{equation}
The first constraint corresponds to the vehicle dynamics given by equation (\ref{bicycle_dynamics_equation}). The second and third constraints impose the bounds on the control inputs and system state variables, which are derived from the physical properties of the chosen vehicle model.

\subsection{Neural Network and Safety Controller Integration}
The output of the neural network is expressed in the ego vehicle coordinate system, whereas the physics-informed controller requires an input that is expressed in the global map coordinate system. Therefore, a coordinate transformation needs to be performed for the waypoints and the obstacles. The traffic information includes the probability of encountering a red light, a stop junction, and being on-road for the agent. Braking will be applied to the agent if the red light probability is above a certain threshold and the throttle will be disabled. Stop junction probability helps to agent to slow down during junction. On-road probability $P_{on\_road}$ tells if agent is on the road or at junction. $P_{on\_road}$ is used to vary the gain of obstacle potential field and weight position and yaw angle tracking terms in the cost function.
\begin{equation}\label{EQ7}
    K_{o} = \dfrac{K}{P_{on\_road} + 0.5}
\end{equation}
\begin{equation}\label{EQ8}
    w_{p_x, p_y, \varphi} = w*(P_{on\_road} + 0.5)
\end{equation}
where $w_{p_x, p_y, \varphi}$ is the weight of position, yaw angle following. $P_{on\_road}$ has value close to $0$ when neural network predicts that agent is not on-road, which leads to equation~\ref{EQ7} to amplify the potential function gain $K_{o}$ and equation~\ref{EQ8} to reduce the weight of position and yaw angle tracking. The above implementation results in MPC controller not precisely following the predicted trajectory and allowing overtaking actions when there is close obstacles at junction.

The overall design is therefore able to 1) precisely track the trajectory for most of the time and 2) perform its own decision-making of overtaking or obstacle avoidance when neural network generated trajectory is not optimal.

\begin{figure*}[t]
\centering
\subfigure[Y Junction]{
\includegraphics[width=0.315\linewidth]{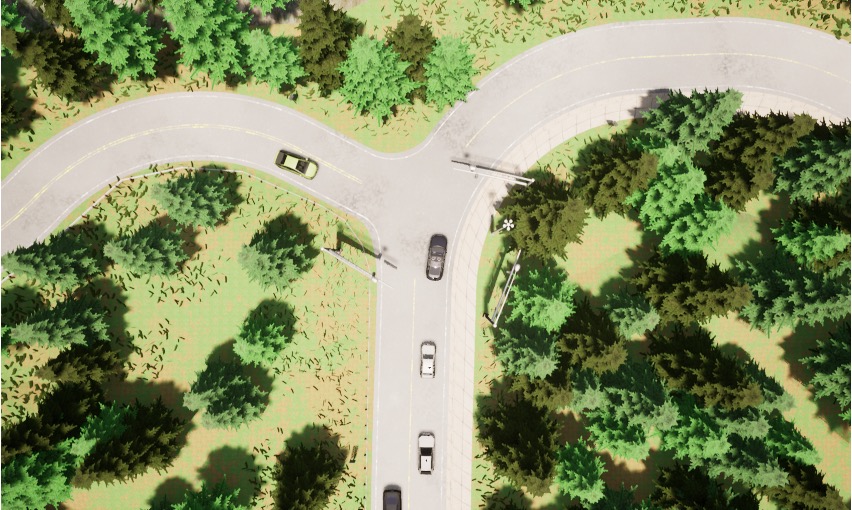}
\label{fig:4-1}
}\hspace*{-0.5em}
\subfigure[Roundabout]{
\includegraphics[width=0.315\linewidth]{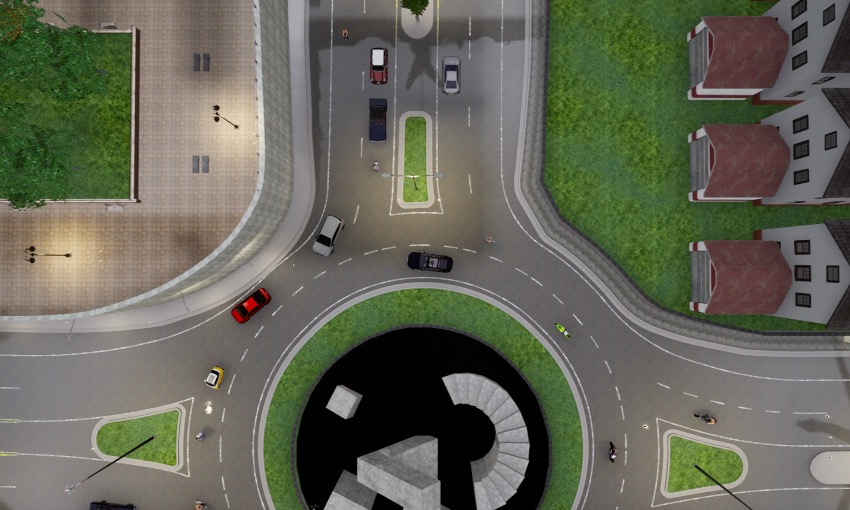}
\label{fig:4-2}
}\hspace*{-0.5em}
\subfigure[Highway Exit]{
\includegraphics[width=0.315\linewidth]{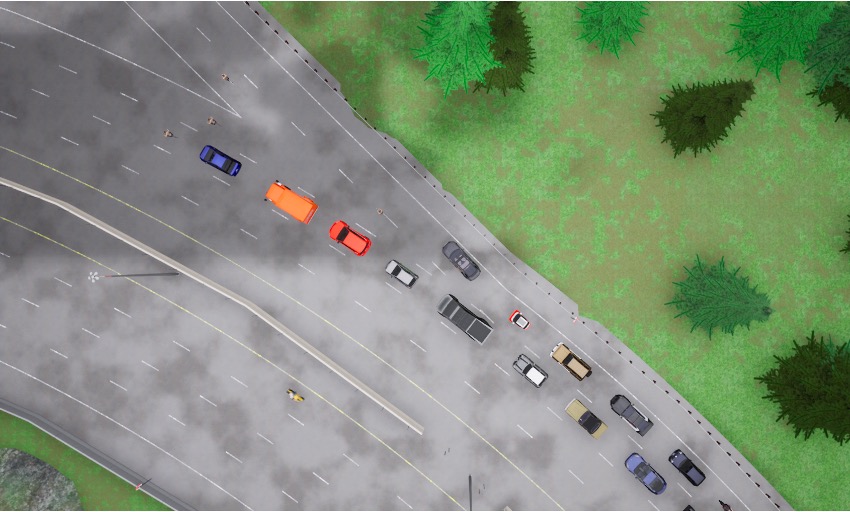}
\label{fig:4-3}
}\\[-0.4ex]

\subfigure[T Junction]{
\includegraphics[width=0.315\linewidth]{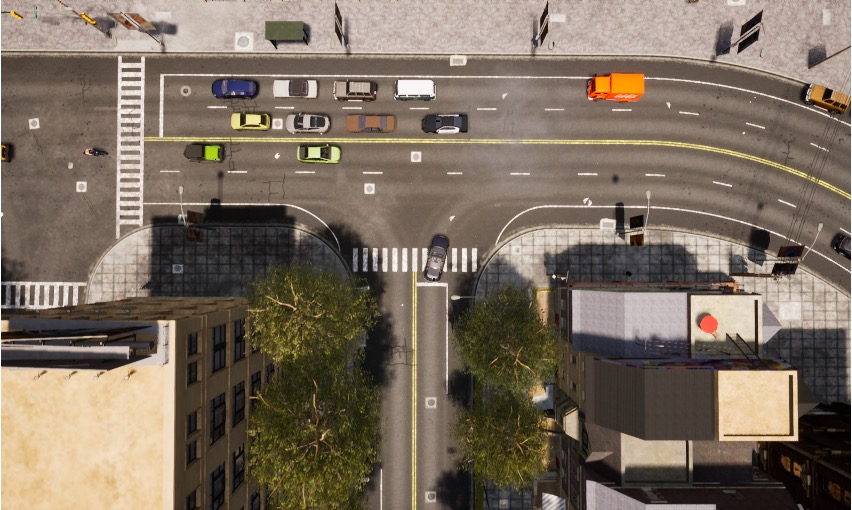}
\label{fig:4-4}
}\hspace*{-0.5em}
\subfigure[Parking Space]{
\includegraphics[width=0.315\linewidth]{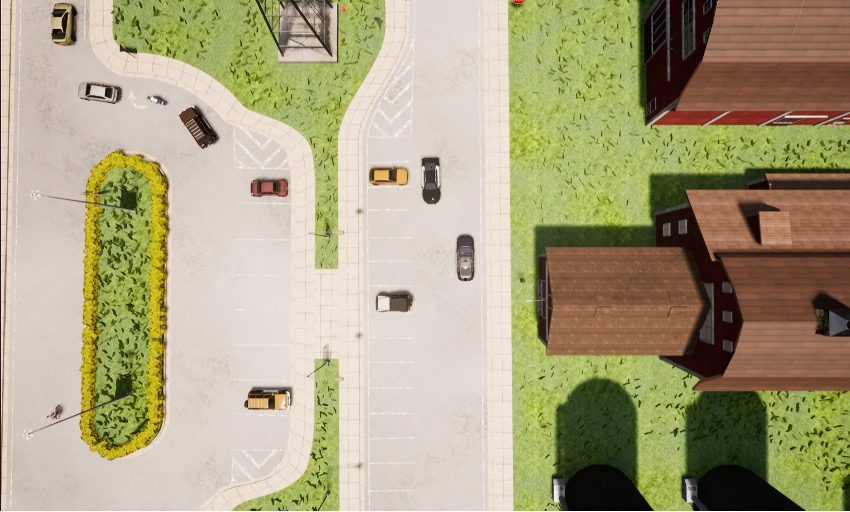}
\label{fig:4-5}
}\hspace*{-0.5em}
\subfigure[Reckless Pedestrian Crossing]{
\includegraphics[width=0.315\linewidth]{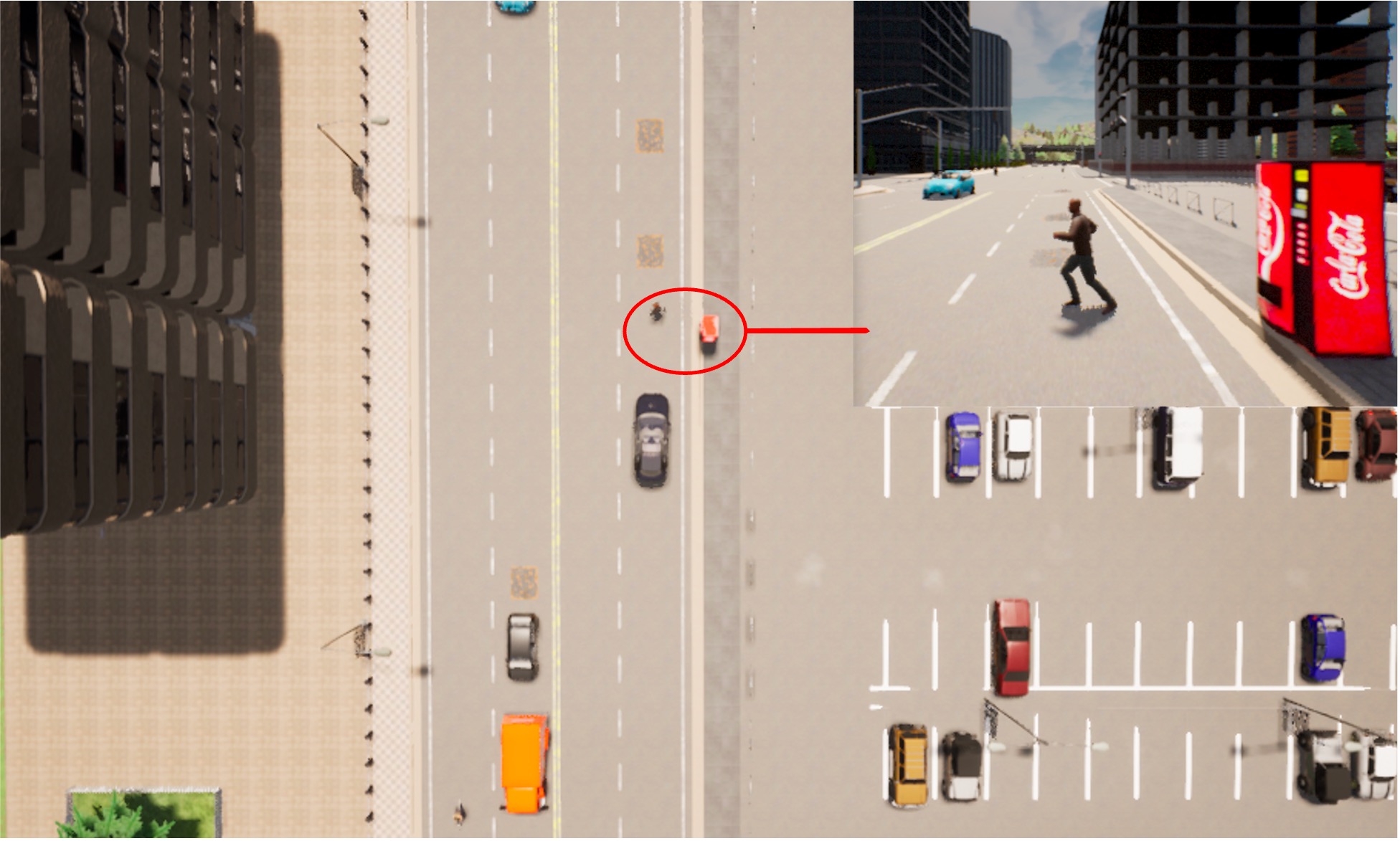}
\label{fig:4-6}
}
\caption{Test Examples of FusionAssurance completing various scenarios under different wearther conditions.}
\label{fig:4}
\end{figure*}

\section{Experiments}
This section presents a comprehensive overview of the experimental setup, encompassing data collection, benchmarking and metrics for evaluation. The proposed method undergoes rigorous testing across a spectrum of intricate and dynamic scenarios which include those depicted in Fig.~\ref{fig:4}. Furthermore, Section \ref{cornercase} aims to demonstrate the efficacy of the proposed methods in selected corner cases where former end-to-end driving algorithms encounter problems. 
\subsection{Environment, Data Collection and Benchmark}
 Autonomous driving framework and experiments are conducted on the open-source simulator CARLA version 0.9.10.1. A dataset of 1M frames was collected from an agent that followed rule-based policy with access to the privileged information of CARLA as groundtruth. Data was collected on all available 8 towns and 21 weather conditions at 2 Hz rate. Routes, dynamic objects and adversarial scenarios were randomly generated as provided in~\cite{shao2022interfuser} to increase the diversity of the collected data. Each frame of dataset contain information of Front, Left and Right Camera Images, Lidar Point Cloud, GPS coordinates and corresponding ground-truth object detection labels generated by simulator.

Evaluations and experiments were conducted on CARLA 42 routes benchmark as its convenience of visualization and variety of scenarios. MPC controller uses optimiser heavily and its performance  highly depends on the CPU power, thus visualization is important for hyperparameter tuning. The benchmark requires the intelligent agent to follow predefined routes without colliding or breaking traffic rules in the presence of adversarial events. The benchmark randomly generates target points and a list of goal locations in global map coordinates for each run. The proposed method utilizes these goal locations to guide the agent to drive without manually setting high-level navigational commands. Three metrics are introduced by the CARLA Leaderboard~\cite{dosovitskiy2017carla} to evaluate our method, which are the route completion ratio (RC), infraction score (IS) and the driving score (DS). The driving score is the product of the infraction score and route completion ratio. It is overall a more comprehensive metric to measure agent planning and safety control capability.


\subsection{Effectiveness Analysis on Corner Cases}
The CARLA leaderboard algorithms have demonstrated their performance on vast majority of scenarios. However, there are still some corner cases that previous work failed to pass. Our proposed method completes complex scenarios (Figure~\ref{fig:4}) and corner cases that previous algorithms could not handle. This section presents several corner cases that the previous state-of-the-art algorithms failed to pass, and illustrates the benefit of our framework. Some of these corner cases occurred to the rule-based agent during data collection, and were consequently learned by the neural agents that adopted a similar policy.\label{cornercase}

\begin{figure}
\centering
\includegraphics[ width=1.0\linewidth]{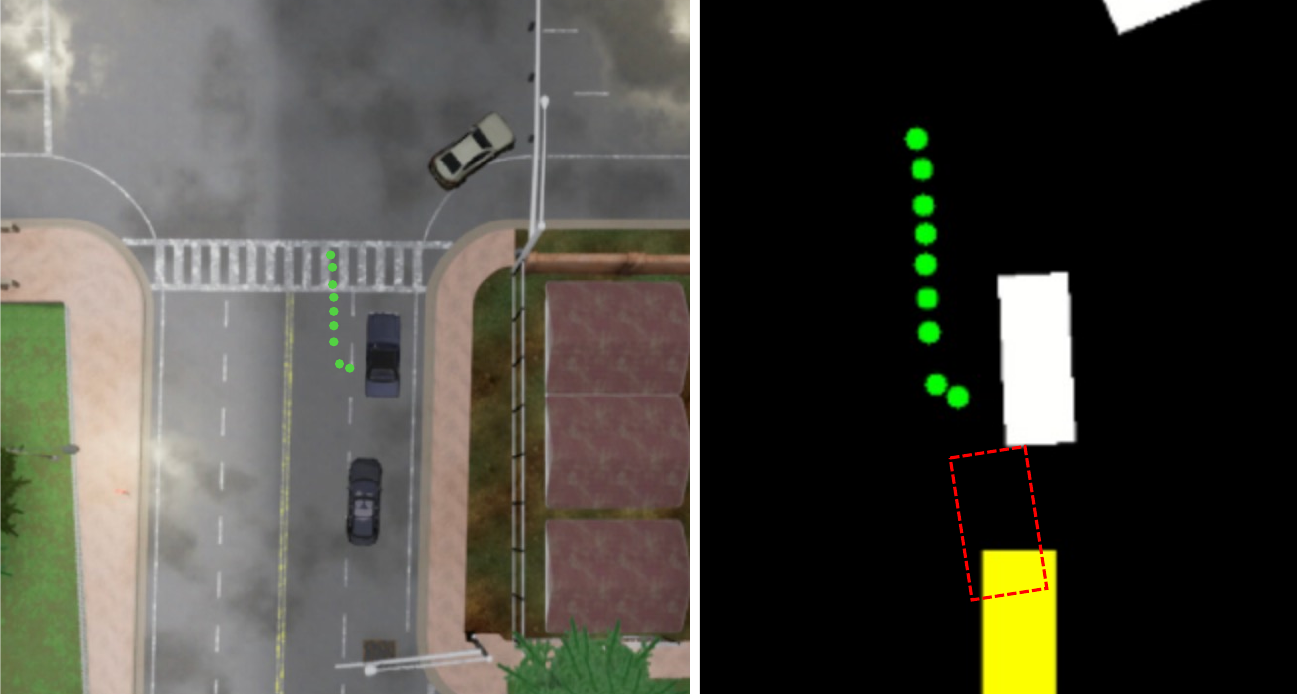}\\[-1ex]
\caption{Case when unfeasible trajectory is generated by neural network.}
\label{fig:bad_trajectory}
\end{figure}

\subsubsection{Case 1(Fig.~\ref{fig:bad_trajectory})} This case shows an anomalous predicted trajectory and it occurs when the neural network is not able to generalize for specific kind of scenarios. It can be seen from the figure that the first available waypoint is 4 meters in front of the ego vehicle and the target vehicle is also $3.5m$ in the front. Collision might occur for InterFuser's safety controller as the target vehicle is not on ego vehicle's trajectory and agent would not slow down. The InterFuser's safety controller simpily is not aware of the existence of the target vehcile under such corner case. Interfuser would directly drives towards the nearest waypoint and red-dash bounding box is where the collision might happen. Our proposed physics-informed safety controller employs Potential Functions and Model Predictive Control to generate a smooth and collision-free trajectory for the agent, which allows the agent to navigate safely through the scenario.

\begin{figure}[t]
\centering
\subfigure[no physics-informed controller]{
\includegraphics[width=0.49\linewidth]{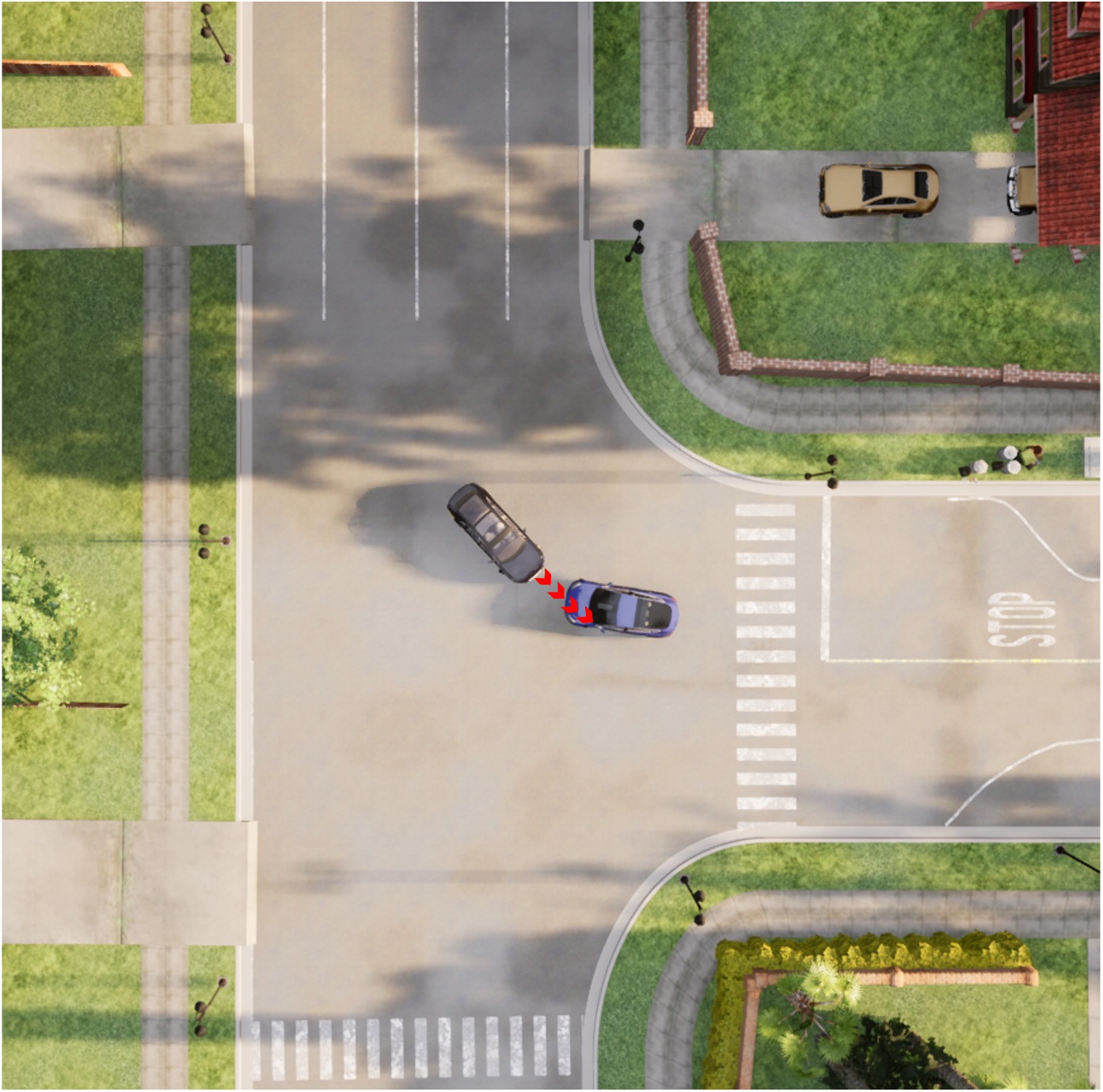}
\label{fig:f5-1}
}\hspace*{-0.7em}
\subfigure[with physics-informed controller]{
\includegraphics[width=0.49\linewidth]{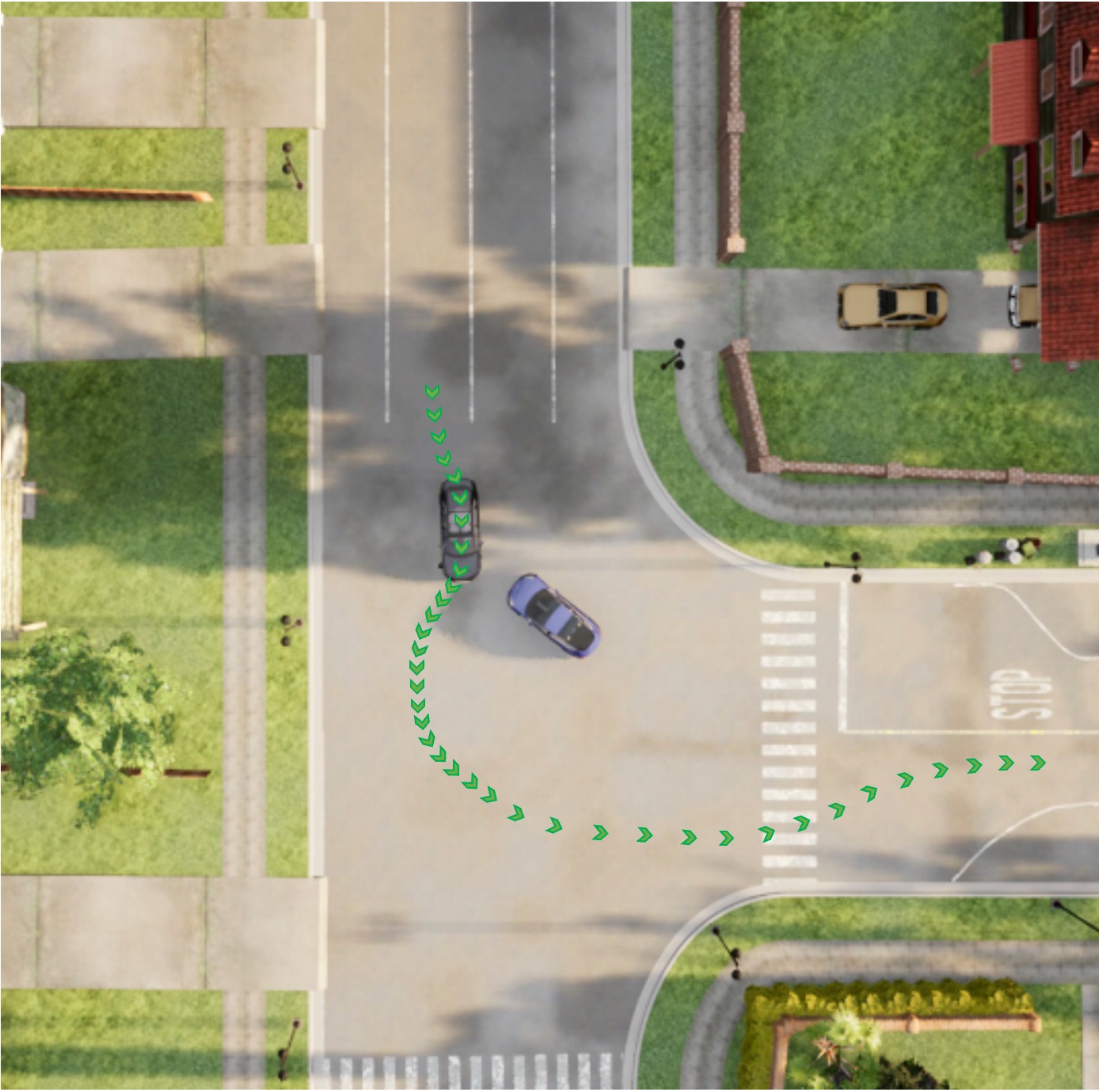}\label{fig:f5-2}
}
\\[-1ex]
\caption{The left figure shows the agent get into Deadlock situation by the neural network planner. The right figure shows the guided trajectory of FusionAssurance under the same planning network.}
\label{fig:trajectory}
\end{figure}

\begin{figure}[t]
\centering
\includegraphics[width=1.0\linewidth]{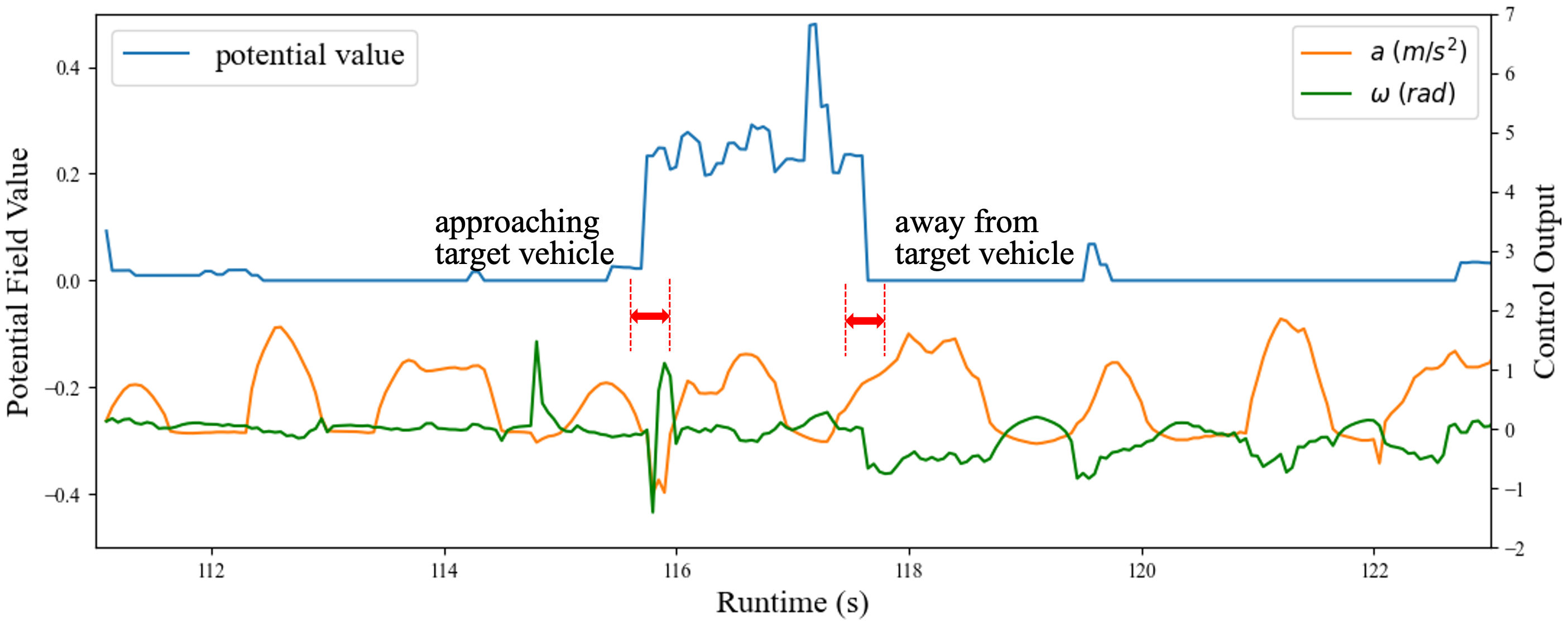}\\[-1ex]
\caption{Obstacles Potential Field value and control output for Case 2.}
\label{fig:pf_control}
\end{figure}

\subsubsection{Case 2(Fig.~\ref{fig:trajectory})} This case is at an T-junction where ego vehicle is driving into an intersection and oncoming cars have the right of way. Despite the two vehicles sharing different paths ideally, the rule-based target agent and trained ego agent drive towards each other as the target vehicle takes a wide turn and the ego vehicle takes a short turn. Two vehicles result in a deadlock situation where two agents stop and give way to each other but none of them are intelligent enough to overtake another.

This is a typical case caused by a low-quality training dataset where the recorded data perform a bad example for the neural network. Our proposed physics-informed safety controller overcomes this issue with the designed obstacle potential field function and performs its own overtaking decision. The physics-informed controller detects the potential field of the vehicle from a far distance and incorporates it in its receding horizon optimization. In Fig.~\ref{fig:pf_control}, when the target vehicle approaches the ego vehicle at time $115.5$s, the obstacle potential value start to rise. the ego vehicle rapidly decelerates and steers away from the target vehicle. The ego vehicle continuously plans new trajectory and returns to the planned trajectory as the target vehicle diminishes at time $117.5$s. The ego vehicle thus manages to execute overtaking decision and control by taking a wider turn than the original predicted trajectory.

\begin{figure}[t]
\centering
\subfigure[MPC$(w_{p_x,p_y}\! \!=\!\!15)$ ]{
\includegraphics[width=0.49\linewidth]{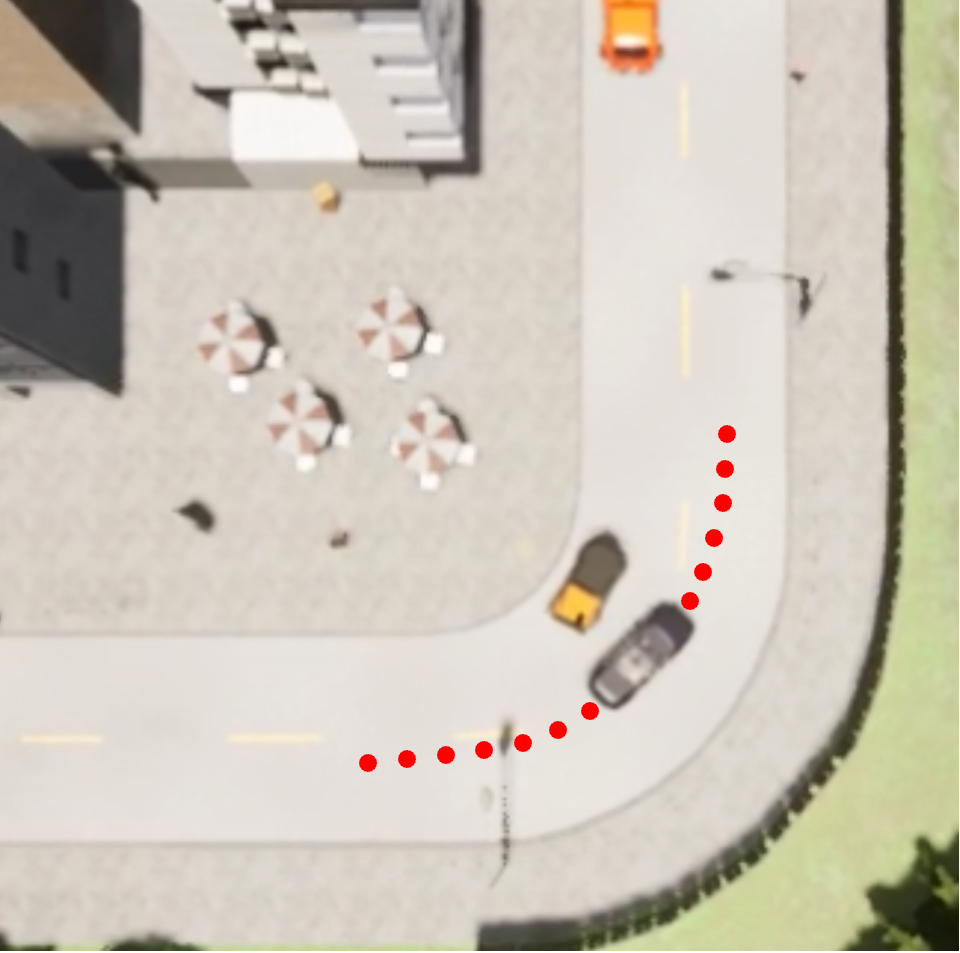}
\label{fig:5-1}
}\hspace*{-0.7em}
\subfigure[MPC$(w_{p_x,p_y}\! \!=\!\!15)$+PF$(K_o\!\!=\!\!60)$]{
\includegraphics[width=0.49\linewidth]{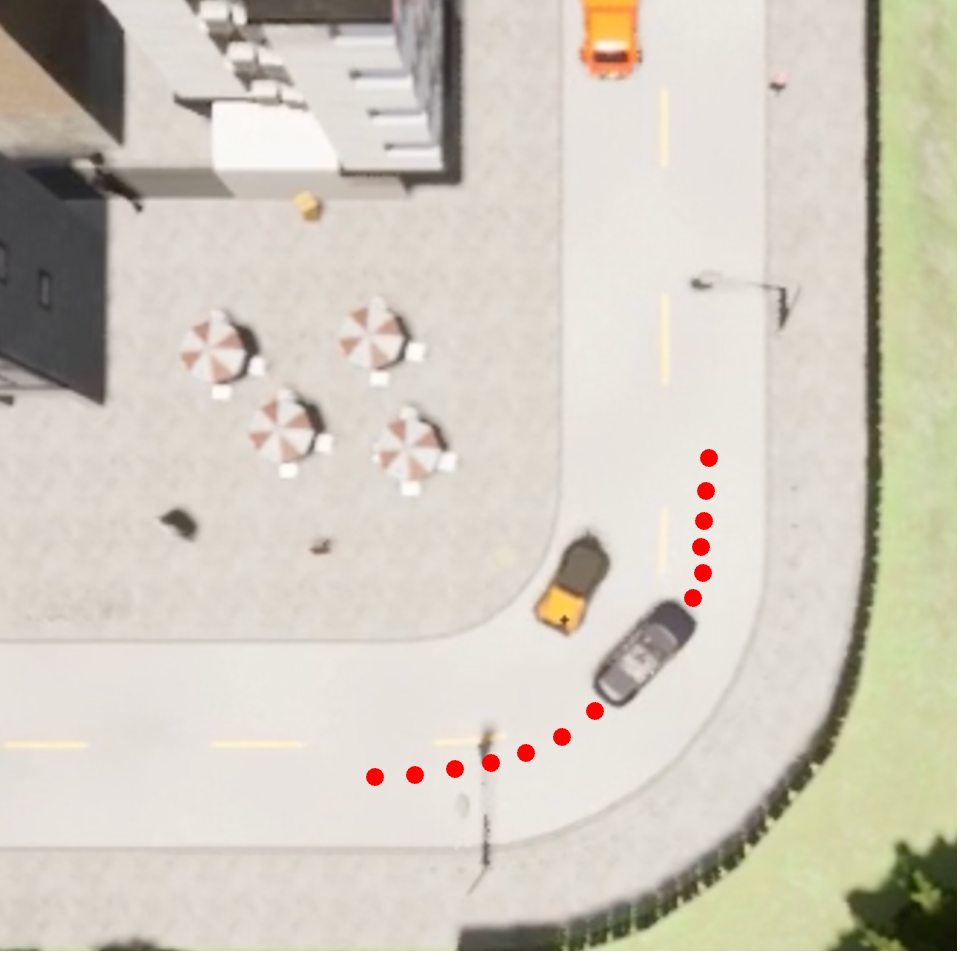}\label{fig:5-2}
}
\\[-1ex]

\caption{The trajectory of the agent with different parameters. (a) and (b) shows the waypoints of agent using MPC and agent using MPC with PF.
}
\label{fig:mpcVSmpc_pf}
\end{figure}

\begin{figure}[t]
\centering
\includegraphics[ width=0.8\linewidth]{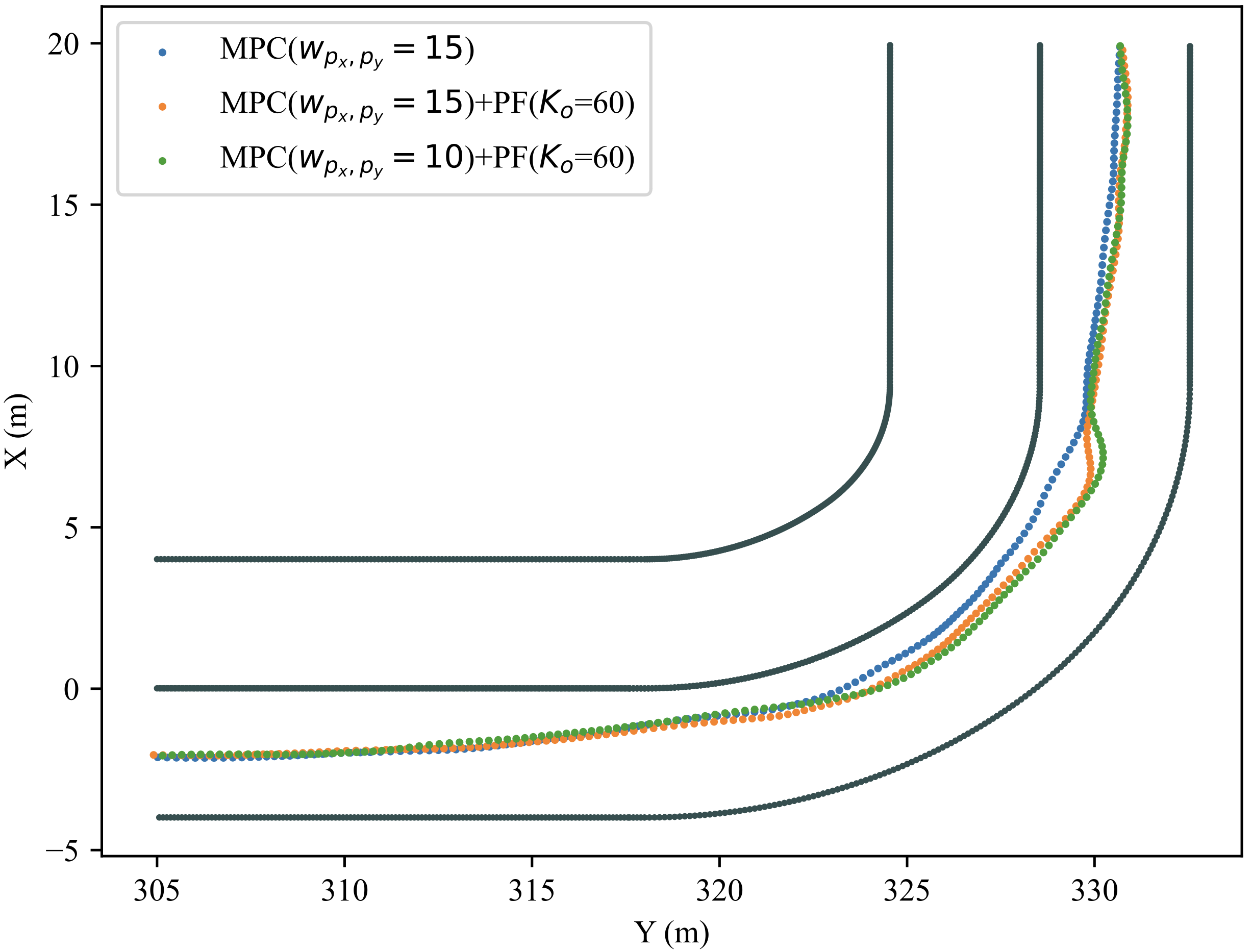}\\[-1.5ex]
\caption{Trajectory comparison of three sets of parameters under Case 3.}
\label{fig:parameter_tuning}
\end{figure}


\subsubsection{Case 3(Fig.~\ref{fig:mpcVSmpc_pf})} 
This case aims to analysis the effect of different safety controller parameters to the driving behavior. The case happens when agent is taking a corner at single carriageway. The MPC controller exhibits a corner-cutting behavior where a portion of the ego vehicle crosses into the opposite lane while following the trajectory. This behavior poses a safety risk if there is an oncoming target vehicle that makes a large turn. 

With the assistance of Potential Field Function, the agent keeps the target vehicle's position in motion consideration. As shown in Figure.~\ref{fig:parameter_tuning}, agent with Potential Field Function has $0.6m$ larger safety distance when it is closest to the target vehicle. By lowering the position tracking parameter $w_{p_x, p_y}$ to 10, agent has further $0.1m$ safety distance. Users can also adjust the agent's driving behavior by tuning the values of target speed, weights for the position tracking $w_{p_x, p_y}$ and gain of obstacle potential function $K_o$. In general, smaller value of $w_{p_x, p_y}$ and higher value of $K_o$ tends to lead to more overtaking actions.

\subsection{Comparison with State of the Art}
The evaluations are done on the CARLA 42 routes benchmark. CARLA 42 routes benchmark encompasses a diverse set of 42 routes across 6 distinct towns. This comprehensive testing suite combines 7 weather patterns with 6 daylight scenarios to ensure each route presents a unique driving experience.. We take the state-of-the-art algorithms on CARLA online leaderboard - \textbf{InterFuser}~\cite{shao2022interfuser}, \textbf{TransFuser}~\cite{Chitta2022PAMI,Prakash2021CVPR}, \textbf{TransFuser++}~\cite{Jaeger2023ICCV}  and \textbf{TCP}~\cite{wu2022trajectoryguided} for comparison. The CARLA benchmark imposes substantial penalties for infractions and therefore most prior work adopt conservative driving behaviors that prioritize avoiding infractions over completing routes. The conservative driving behaviors often lead to Deadlock situation where stuck agents are not intelligent enough to overtake another.

\textbf{InterFuser(reproduced)} and FusionAssurance use the exact same training data for performance comparison. FusionAssurance outperforms InterFuser by $9.6$\% in driving score with higher route complement ratio $99.966$\% and fewer infractions with $0.907$. \textbf{TransFuser++} has the best infraction score but their driving behavior is conservative. Similar to other previous work, TransFuser++ stops for all potential obstacles without much overtaking. It often gets into Deadlock situation and its route complement ratio is $9$\% less than our proposed method. Our proposed method has the highest driving score and route complement ratio. This is mainly due to adaptive cruise control, obstacle avoidance and overtaking capability provided by safety controller. 

\begin{table}
\centering
\caption{Comparison with State-of-the-Art on CARLA 42 routes Benchmark.}
\begin{tabular}{lccc} 
\toprule 
\multirow{2}{1.0cm}{\centering Method}& \multirow{2}{0.9cm}{\centering Route Comple$\uparrow$}& \multirow{2}{1.0cm}{\centering Infraction Rate$\uparrow$}& \multirow{2}{0.8cm}{\centering Driving Score$\uparrow$}\\
\\
\midrule 
InterFuser (reproduced)~\cite{shao2022interfuser} & 93.827 & 0.865 & 81.084\\
Latent Transfuser (pretrained)~\cite{Chitta2022PAMI} &87.359 & 0.756 &63.570  \\
Transfuser++ (pretrained)~\cite{Jaeger2023ICCV} & 91.989 & \textbf{0.982} & 90.203  \\
TCP (pretrained)~\cite{wu2022trajectoryguided}  & 90.946 & 0.912       & 82.991\\
FusionAssurance (ours)  & \textbf{99.966} & 0.907 & \textbf{90.683}\\
\midrule 
\end{tabular}
\text Note: Infraction rate does not consider travelled distance. Driving score is more comprehensive for measuring infraction per unit distance. [\ref{key}]
\end{table}


\subsection{Ablation Study}

The effectiveness of each module is studied in this section. FusionAssurance(histogram) PID uses histogram for lidar feature extraction and  rule-based PID for vehicle motion control. PointPillars preserves more 3D point cloud features compared to the 2d lidar histogram. This helps the network to understand the surroundings, especially the detection of small-sized objects and differentiation of objects with different height. Under same controller setting, pointpillars feature extraction methods have up to $7.5$\% driving score increase. 

With the replacement of PID controller by the MPC controller, the method is able to achieve a higher overall driving score and higher route complement rate both by $1$\%. The MPC controller optimizes not only the nearest waypoints, but also a horizon of future waypoints under the specified vehicle dynamics. This indicates MPC controller exhibits better performance in trajectory tracking and thus achieves a higher route completion ratio.

By replacing the controller from PID to our proposed physics-informed safety controller(MPC+PF), FusionAssurance(histogram) and FusionAssurance(pointpillars) gain increase in driving score by $2.1$\% and $9.2$\% repectively. 
The full version FusionAssurance which deploys pointpillars, MPC and PF has a siginificant performance boost on all three metrics.

\begin{table}
\centering
\caption{Ablation Study on the Network Structure and Safety Controller.}
\begin{tabular}{lccc} 
\toprule 
\multirow{2}{1.0cm}{\centering Method}& \multirow{2}{0.9cm}{\centering Route Comple$\uparrow$}& \multirow{2}{1.0cm}{\centering Infraction Rate$\uparrow$}& \multirow{2}{0.8cm}{\centering Driving Score$\uparrow$}\\
\\
\midrule 
FusionAssur(histogram) PID  & 93.827 & 0.865 & 81.084\\
FusionAssur(histogram)   MPC+PF  & 97.177  & 0.857 & 83.184\\
FusionAssur(pointpillar) PID & 96.970 & 0.840 & 81.435 \\
FusionAssur(pointpillar) MPC & 97.839& 0.844 &82.576 \\
FusionAssur(pointpillar) MPC+PF  & 99.966 & 0.907 & 90.683\\
\midrule 
\end{tabular}
\end{table}

\subsection{Discussions}
The CARLA infraction rate does not take travelled distance into account. Agents which travel short distance and get into Deadlock situation can still get high infraction rate. Driving score which is the product of route complement and infraction rate is therefore a more comprehensive metric for measuring infraction rate per unit distance. Therefore FusionAssurance achieves the best performance on infraction per unit distance.\label{key}

\section{Conclusions}

In this paper, we propose FusionAssurance, an integrated end-to-end autonomous driving framework with safety controller. This framework employs a neural network for perception and trajectory planning, and Model Predictive Control with Potential Field for enhanced safety control. The safety of the neural network planning task is ensured by the low-level control with the assistance of designed Potential Functions. The overall framework also makes integrated overtaking decisions to compensate for the suboptimal network planning results and to circumvent potential deadlock. With same training dataset, FusionAssurance outperforms baseline - InterFuser by $9.6$\% on CARLA 42 routes benchmark. Additionally, the physics-informed safety controller can adapt any trajectory-based end-to-end neural network to boost the overall performance.

Potential future work includes adding lane detection to the neural network and using lane information as Potential Field to reduce the out-of-lane rate. Another future improvement is to develop static obstacle detection ability for the network to further prevent infraction.

\section*{ACKNOWLEDGMENT}

This work is partially supported by National Natural Science Foundation of China under grants 62303389 and 62373289, Guangdong Basic and Applied Basic Research Funding under grants 2024A1515012586 and 2022A151511076, Guangzhou Basic and Applied Basic Research Scheme under grant 2023A04J1067 and Guangzhou Municipality-University Joint Funding under grant 2023A03J0678.

\addtolength{\textheight}{-0cm}   





\bibliographystyle{ieeetr}
\bibliography{refs}
\end{document}